\documentclass[runningheads]{llncs}
\usepackage{graphicx}

\usepackage{tikz}
\usepackage{comment}
\usepackage{amsmath,amssymb} 
\usepackage{color}

\usepackage[accsupp]{axessibility}  

\usepackage{subfig}
\usepackage{graphicx}
\usepackage{booktabs}
\usepackage{microtype}
\usepackage{xspace}
\usepackage{tabularx}
\usepackage{multirow}
\usepackage[outercaption]{sidecap}    
\usepackage{algorithm,algpseudocode}
\usepackage{url}
\usepackage{chngpage}
\usepackage{array}
\usepackage{bbding}
\usepackage{verbatim}
\usepackage{wasysym} 
\usepackage{booktabs}  
\usepackage{xcolor}
\usepackage{color, colortbl}
\definecolor{mygray}{gray}{0.95}
\definecolor{myred}{rgb}{1.0, 0.0, 0.0}

\usepackage[font=footnotesize,labelfont=bf]{caption} 
\usepackage[pagebackref,breaklinks,colorlinks]{hyperref}
\usepackage{threeparttable}
\usepackage{sidecap}

\makeatletter
\DeclareRobustCommand\onedot{\futurelet\@let@token\@onedot}
\def\@onedot{\ifx\@let@token.\else.\null\fi\xspace}

\def\etal{\emph{et al}\onedot}

\makeatother
\newcommand{\midsepchange}{\aboverulesep = 0.15mm \belowrulesep = 0.484mm}
\midsepchange

\begin{document}
\pagestyle{headings}
\mainmatter
\def\ECCVSubNumber{839}  

\title{Less than Few:\\ Self-Shot Video Instance Segmentation}

\titlerunning{Self-shot learning}
%
\author{Pengwan Yang \and Yuki M. Asano \and
Pascal Mettes \and Cees G. M. Snoek}
\authorrunning{P. Yang et al.}
%
\institute{University of Amsterdam\\
\email{yangpengwan2016@gmail.com}}
\maketitle

\begin{abstract} 
The goal of this paper is to bypass the need for labelled examples in few-shot video understanding at run time. 
While proven effective, in many practical video settings even labelling a few examples appears unrealistic. This is especially true as the level of details in spatio-temporal video understanding and with it, the complexity of annotations continues to increase. Rather than performing few-shot learning with a human oracle to provide a few densely labelled support videos, we propose to automatically learn to find appropriate support videos given a query. We call this self-shot learning and we outline a simple self-supervised learning method to generate an embedding space well-suited for unsupervised retrieval of relevant samples. 
To showcase this novel setting, we tackle, for the first time, video instance segmentation in a self-shot (and few-shot) setting, where the goal is to segment instances at the pixel-level across the spatial and temporal domains. We provide strong baseline performances that utilize a novel transformer-based model and show that self-shot learning can even surpass few-shot and can be positively combined for further performance gains. Experiments on new benchmarks show that our approach achieves strong performance, is competitive to oracle support in some settings, scales to large unlabelled video collections, and can be combined in a semi-supervised setting.

\end{abstract}

\section{Introduction}
The goal of this paper is to decrease the reliance on humans to provide labelled examples in few-shot video understanding. While impressive few-shot video classification~\cite{kliper2011one,CaoCVPR20,PerrettCVPR21}, localization~\cite{feng2018video,zhang2020metal,yang2020localizing} and detection~\cite{feng2019spatio,yang2021few} results have been reported, in many practical video settings even labelling a few examples may appear unrealistic. 
This is especially true as the level of spatio-temporal video understanding and with it, the complexity of annotations continues to increase. 
Consider for example the problem of video instance segmentation \cite{yang2019video,bertasius2020classifying,wang2021end}, where datasets for example contain around 1.6K annotated frames for just a single object class. 
We deem it unlikely that an interacting user, 
that is looking to segment a ``query'' video with unknown instances,
is willing to provide pixel-precise annotations masks for all objects in a frame for each video in the support set, despite this being a setting which is typical for more classical, image-based few-shot learning scenarios. 
Thus, rather than relying on a human oracle to provide a few densely labeled support videos, we propose to automatically learn to find appropriate support videos given a query.

For this, we introduce the notion of \textit{self-shot} learning, in which the need for labelled video clips at test-time is abolished. 
Instead, one is provided with a large unlabelled pool of videos from which samples potentially relevant to the query video can be retrieved and utilized in a strictly unsupervised fashion. 
We address this by adapting a simple self-supervised learning method~\cite{varamesh2020self} to generate an embedding space well-suited for unsupervised retrieval of relevant samples. 
To showcase this novel setting, we go beyond just bounding box detection and temporal localization and instead tackle, for the first time, \textit{video instance segmentation} in a self-shot (and few-shot) setting, where the goal is to segment instances at the pixel-level across the spatial and temporal domains.

Overall, we make three contributions in this paper:
\begin{enumerate}
    \item We propose the setting of self-shot learning. While annotations are used during training (similar to few-shot), at test-time, new classes are evaluated \textit{without} any annotations, but with access to an unlabelled dataset.
    \item We investigate this new setting for a particularly annotation-heavy scenario, that of video instance segmentation, for which we propose new splits to establish a self-shot (and few-shot) benchmark.
    \item  Finally, we provide strong baseline performances that utilize a novel transformer-based model and show that self-shot learning can even surpass few-shot and can be positively combined for further performance gains.
\end{enumerate}

\section{Related work}

\paragraph{Video few-shot learning.} 
There is limited related work on the few-shot learning setup for videos. Initial works have explored few-shot learning for the task of video classification~\cite{kliper2011one,zhu2018compound}. For example, OSS-Metric Learning~\cite{kliper2011one} measures similarity of pairs of video to enable few-shot video classification. Yang~\etal~\cite{yang2018one} introduce few-shot action localization in time, where a few positive labelled and several negative labelled support videos steer the localization via an end-to-end meta-learning strategy.
%
%
%
Xu~\etal~\cite{xu2020revisiting} and Zhang~\etal~\cite{zhang2020metal} also perform few-shot temporal action localization with the assistance of video-level class annotations. 
To further free the need for labels, a new research line is emerging, called few-shot common action localization, where the common action in a long untrimmed query video is localized in time~\cite{feng2018video,yang2020localizing} or both in time and in space~\cite{feng2019spatio,yang2021few} based on a few support videos containing the same action. 
Closest to our work is few-shot spatio-temporal action localization by Yang~\etal~\cite{yang2021few}, who adopt a transformer-based action detection architecture and extend to localizing actions at pixel level. 
They propose a mask head upon the action detection boxes to perform the binary classification for the pixels inside each detected box. 
In this paper, we propose the new task of self-shot video instance segmentation that operates on objects instead of actions, predicts the segmentation directly, removes the need for having predefined (labelled) support videos, and encapsulates the few-shot setting as a special case. 


\paragraph{Video zero-shot learning.} 
Various video tasks have been investigated from a zero-shot perspective. Zhang~\etal~\cite{zhang2020zstad} can localize the unseen activities in time in a long untrimmed video based on the label embeddings. 
Spatio-temporal action localization is also explored in zero-shot setting by linking actions to relevant objects~\cite{jain2015objects2action,mettes2017spatial,mettesIJCV21}, or by leveraging trimmed videos used for action classification~\cite{jain2020actionbytes}. 
Wang~\etal~\cite{wang2019zero} achieve zero-shot video object segmentation by proposing a novel attentive graph neural network which can iteratively fuse information over video graphs. 
Dave~\etal~\cite{dave2019towards} can segment the moving objects in videos, even ones unseen in training. 
Just like the zero-shot setting, we also aim to segment unseen object instances in videos, without any labelled support videos. 
But we try to leverage self-retrieved (free) support videos to boost the performance.

\paragraph{Self-supervised learning.} 
Self-supervision has been proposed as a method to obtain feature representations without labels.
This has been accomplished by geometric pretext tasks~\cite{gidaris2018unsupervised,noroozi2016unsupervised,pathakCVPR16context}, clustering~\cite{asano20self-labelling,caron18deep,caron20unsupervised,gidaris2020learning} or more recently contrastive~\cite{kalantidis2020hard,chen2020exploring,misra2020pirl,chen20a-simple,wu2018unsupervised} and teacher-student approaches~\cite{caron2021emerging,grill2020bootstrap}.
These have also been extended to the video domain~\cite{feichtenhofer2021large,patrick2021space,patrick2021compositions,qian2021spatiotemporal,korbar2018cooperative,asano2020labelling,alwassel2020self,benaim2020speednet,han2020self,kim2019self}.
In this work we use self-supervised learning to construct an embedding space well-suited for retrieving semantically relevant videos to support video instance segmentation. 
Note that this use of support samples is fundamentally different to how it has been used in other self-supervised works such as~\cite{dwibedi2021little} or \cite{patrick2020support}, where they are only used as random subsets to aid contrastive learning.
Instead, support samples are the goal of our self-shot method.
To this end, we compare the use of noise contrastive methods~\cite{wu2018unsupervised} and the differentiable ranking loss~\cite{varamesh2020self} extended to the video domain.

\paragraph{Video instance segmentation (VIS).}
VIS~\cite{yang2019video,bertasius2020classifying} requires classifying, segmenting, and tracking instances over all frames in a given video. 
With the introduction of the YouTube-VIS dataset, which contains dense pixel-level annotations across consecutive frames~\cite{yang2019video}, considerable progress has been made in tackling this challenging task. 
State-of-the-art methods typically develop sophisticated pipelines and rely on heavy supervision and complex heuristic rules to associate the instances across frames. 
As two representative methods, MaskTrack R-CNN~\cite{yang2019video} extends the Mask R-CNN model~\cite{he2017mask} with a pair-wise identity branch to solve the instance association problem in VIS, while MaskProp~\cite{bertasius2020classifying} introduces a multi-stage framework~\cite{chen2019hybrid} for propagating instance masks in time. 
In contrast, VisTR~\cite{wang2021end} builds a DETR-based pipeline~\cite{carion2020detr} for the VIS task in a query-based end-to-end fashion, which can supervise and segment the instances across frames as sequences. 
In this paper we adopt the spirit of VisTR to treat instance segmentation in a query video as a sequence prediction problem. 
Different from the usual VIS task, our self-support video instance segmentation task focuses on recognizing, segmenting, and tracking the instances in a query video containing novel classes -- from just a few retrieved support videos and without knowing any annotations.

\section{Problem definition and benchmarking}
\subsection{Task definition}
Our goal is video instance segmentation in a query video without having access to any labelled training examples with the same instances as in the query. Instead, we consider a self-shot scenario where we have an unlabelled pool of videos that we can use to help guide the instance segmentation. To that end, we
denote a set of seen classes as $\mathcal{S}$ and a disjoint set of unseen classes as $\mathcal{U}$, where $\mathcal{S} \cap \mathcal{U} = \emptyset $. 
Let $\mathcal{D}_{\mathcal{S}}=\{(x,y)|x\in\mathcal{X},y\in\mathcal{Y}^\mathcal{S}\}$ represent the set of labelled training data on seen classes, where $x$ is the pixel-wise feature embeddings from the visual space $\mathcal{X}$, $y$ is the corresponding pixel-wise label in the label space $\mathcal{Y}^\mathcal{S}$ of seen classes. $\mathcal{D}_{\mathcal{U}}$ denotes the set of unlabelled videos on unseen classes. Our self-shot learning shares with few-shot and zero-shot learning the same goal to learn a model from $\mathcal{D}_{\mathcal{S}}$ and predict the label of each pixel for videos in $\mathcal{D}_{\mathcal{U}}$. However, they differ in their objective and expected data availability:

\paragraph{Few-shot learning.}
For each unseen class $c \in \mathcal{U}$, a handful of predefined support videos $\mathcal{V}_c^k$ are provided, where $k$ is small. Then for each query video $Q_c \in \mathcal{D}_\mathcal{U}$, the small set of support videos containing exactly the same class $\mathcal{V}_c^k$ function as guidance videos to predict a segmentation for the unseen object class $c$. 

\paragraph{Zero-shot learning.}
In the most conventional zero-shot strategy~\cite{zhang2020zstad,jain2015objects2action,mettesIJCV21,li2020consistent}, all class labels $\mathcal{C} {=} \mathcal{S}\cup\mathcal{U}$ are provided and mapped through semantic embeddings to vector representations $\{v_c|c\in\mathcal{C}\}$. Then a joint visual-text perspective helps the model learned on the seen classes generalize to the unseen classes.

\paragraph{Self-shot learning.}
Instead of predefined supports or semantic class labels, self-shot learning relies on an unsupervised manner to retrieve support videos $\{ \mathcal{V}_{Q_c}| \mathcal{V}_{Q_c} \in \mathbb{S} \}$ for each query video $Q_c, c \in \mathcal{U}$ from a collection of unlabelled videos $\mathbb{S}$. 
It leverages the discovered support videos as guidance for predicting an instance segmentation. 
Self-shot learning can be viewed as a framework to obtain noisy few-shot examples without the need for human annotations.

\subsection{Datasets}
Since self-shot video instance segmentation is a new task, we set up two benchmarks through the reorganization of two existing (many-shot) video instance segmentation datasets, namely YouTube-VIS~\cite{yang2019video} (2021 version) and OVIS~\cite{qi2021occluded}.

\textbf{Self-VIS.} 
YouTube-VIS contains $2{,}985$ videos in the training set where the instance mask annotations are publicly available. The annotated instances cover $40$ instance categories and a minority of the videos have instances of more than one classes. To build a setting with videos containing one singe instance class, we discard videos with multiple instance classes and obtain a total of $2{,}123$ videos. We randomly select $30$ classes for training and $10$ classes for validation and testing.

\textbf{Self-OVIS.} 
Occluded VIS (OVIS) provides $607$ videos with annotated instance masks. Among the $25$ instance categories in OVIS, $17$ are for training and $8$ for validation and testing. 
With more instances of multiple classes per video, and more frequent occlusions, the setting of OVIS is much harder than the one of YouTue-VIS.
More details are provided in Table~\ref{appx:tab:dataset} in the Appendix.
During training, the query video and support videos are randomly paired according to the common instances present, while the pairs are fixed for validation and testing for reproducibility. 
All dataset construction details will be released for researchers to reproduce and build upon this work. 

\textbf{YouTube-8M Segments.} The YouTube-8M Segments dataset is a subset of the YouTube-8M dataset proposed in the same paper~\cite{abu2016youtube}. 
It contains about 237K 5-second videos extracted from around 50K source videos and while it contains annotations, we do not use any of the labels.
Instead, we adopt YouTube-8M Segments as our unlabelled video database $\mathbb{S}$ for self-shot retrieval and call the self-shot benchmarks

\section{Finding support videos through self-shot learning}
The purpose of self-shot learning is to retrieve videos from a large, unlabelled video dataset that will aid in performing inference on the query video, specifically for the task of instance segmentation in this paper. To this end, we train an encoder that will yield an embedding space well suited for video retrieval by adopting components of self-supervised representation learning methods MoCo~v1 to v3~\cite{he2020momentum,chen2020improved,chen2021empirical}, multiple-instance NCE~\cite{miech2020end} and self-supervised ranking~\cite{varamesh2020self}.

For self-shot learning, we are given an unlabelled video collection $\mathbb{S}$. 
Each clip is encoded by two visual encoders, $\Phi$ and $\tilde{\Phi}$, 
where
$\tilde{\Phi}$ is updated as the exponential moving-average of $\Phi$ as in~\cite{he2020momentum,chen2021empirical}. 
With this setup we evaluate self-shot retrieval with two different losses: noise-contrastive instance discrimination and ranking.

The contrastive loss $\mathcal{L}_{\mathrm{NCE}}$ in our case is given  by setting positive pairs to be different temporal crops of a single video, while negative pairs are constructed from other instances of the dataset.
Let $V_i \in \mathbb{S}$ denote a single unlabelled video and let $v \in V_i$ denote one of its temporal crops.
Then we naturally arrive at the following multiple-instance NCE~\cite{miech2020end}  formulation:
\begin{equation}
\mathcal{L}_{\mathrm{NCE}}(v) = -\log\frac{\sum_{{v}^+}\exp
\langle \Phi{(v)}\cdot\tilde{\Phi}(v^+) \rangle_\tau}{\sum_{({v}^+ \cup {v}^-)}\exp \langle \Phi{(v)}\cdot\tilde{\Phi}(v^+) \rangle_\tau)},
\label{equ:nce_loss}
\end{equation}
where $\langle \cdot, \cdot\rangle_\tau$, is a temperature-scaled dot-product, and $v^+$ is the positive set defined as $V_i \setminus v$ and $v^-$ is the negative set, corresponding to crops from other videos in $\mathbb{S}$.

We further experiment with transplanting a differentiable ranking loss from~\cite{varamesh2020self}, to the setting of 
using two encoders.
The ranking loss is a less aggressive form of enforcing self-invariance than the NCE loss, and is given by learning an embedding space in which a set of positive videos is ranked \textit{above} a set of negative videos, when comparing distances in feature space.
More precisely,
\begin{equation}
\mathcal{L}_{\mathrm{Rank}}(v) = 
-\log\sum_{
{v}^+ }
\frac{R_{\Phi(v)}(\tilde{\Phi}(v),\tilde{\Phi}(v^+))}
{R_{\Phi(v)}(\tilde{\Phi}(v),\tilde{\Phi}({v}^+)\cup\tilde{\Phi}({v}^-))},
\label{equ:rank_loss}
\end{equation}
where $R_{a}(b,c)$ is a differentiable function to rank video $b$ among all videos in the set \{$c$\} with respect to the query video $a$~\cite{brown2020smooth,varamesh2020self}.

Once the feature spaces are learned, the final step is \textbf{retrieving relevant support videos using only the query itself}.
For query video $q$, we use the self-supervised trained encoder $\Phi$  and a simple $k$-nearest neighbor ($k$NN) approach.
The self-shot support videos for query $q$ is obtained as:
\begin{equation}
\text{self-shot}(q)  = k\mathrm{NN}(\Phi(q, \Phi(\mathbb{S}))).
\label{support}
\end{equation}
Note that this simple $k$NN approach allows us to use self-shot learning as a \textit{plug-and-play} component, which can be used to replace supervised support videos or to extend these in a semi-supervised manner. This setup is generally applicable, we focus on video instance segmentation as testbed in this paper, due to the hefty annotation demand for supervised settings.

\textbf{Implementation details.} 
We follow the uniform frame sampling method in~\cite{arnab2021vivit} for mapping a video to a sequence of tokens of a Vision Transformer of size \textit{B} (ViT-B). 
In each mini-batch, we use 160 $5{\times}224{\times}224$ video segments from 32 videos. 
The patch-size of ViT-B is $16{\times}16$. 
We keep the memory queue~\cite{he2020momentum} and the length is 1280. We train for 40 epochs with the AdamW optimizer~\cite{loshchilov2017decoupled}, with an initial learning rating of $10^{-4}$, which we decay by 10 at epoch 25. Further details are provided in the Appendix.

\section{A Self-Shot and Few-Shot VIS Transformer}



Equipped with self-shot learning, we examine a research problem with a high annotation cost: video instance segmentation. Hence our goal is to segment and track the object instances of interest in a query video. We do however not assume access to training videos labelled with the same instance classes, temporal boundaries, or mask annotations as support for the query video at test-time. In essence, the query video is on its own and we are instead given a collection of unlabelled videos. We seek to find a few unlabelled support videos from this collection through self-shot learning. While we aim for self-shot learning, few-shot video instance segmentation using a few support videos has not yet been investigated either. Hence, we first introduce a baseline sequence-to-sequence transformer approach to solve video instance segmentation given semantically similar (support) videos.
At the core of our approach is a common instance segmentation transformer, which contains three stages with three functions: (\textit{i}) encoding and extracting of features for query and support videos, (\textit{ii}) learning of pixel-level similarities for the query features by leveraging cross-attention, and (\textit{iii}) prediction of instance mask sequences across space and time through decoding. Each step is detailed below.
\paragraph{Feature extraction.}
We adopt a modified ResNet-50~\cite{he2016deep} with a bigger receptive field for feature extraction, with the complete architecture given in the Appendix. We let a single query video and a few support videos go through the backbone to extract the pixel-level image frame feature sequences. 
Assume that the query video contains $T$ frames and the support videos contain $T'$ frames in total. 
The backbone generates a lower-resolution activation map for each frame in the query video and support videos, then the frame features are concatenated to form video clip level features for the query and support videos. 
The query video feature is denoted as $f_q \in \mathbb{R}^{T {\times}d {\times} W {\times} H}$ and the support feature is $f_s \in \mathbb{R}^{T' {\times} d {\times} W {\times} H}$. 
The weights of the backbone are shared between the query and support videos.

\begin{figure}[t]
	\centering
	\includegraphics[width=0.99\linewidth]{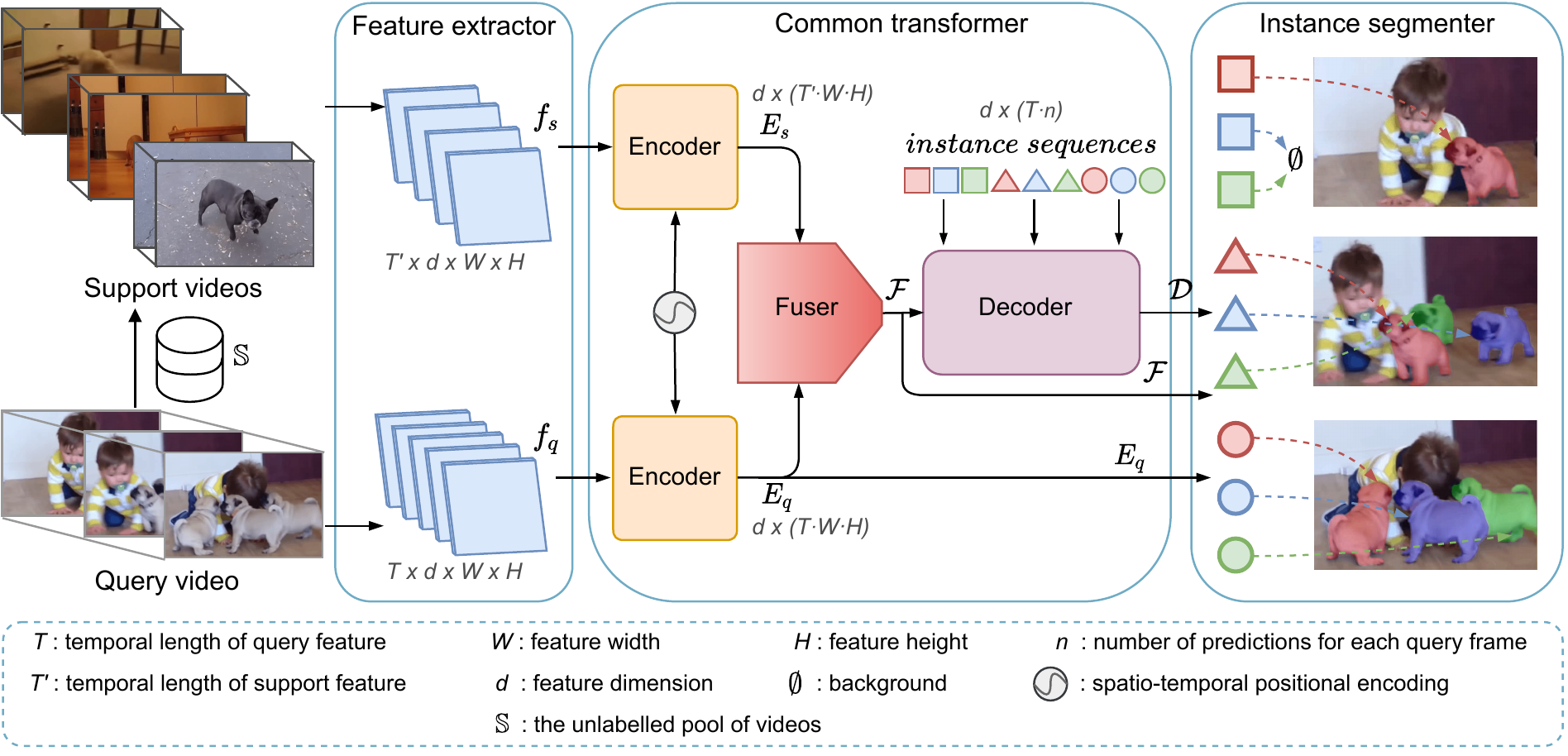}
	\caption{\textbf{Overview of the self-shot and few-shot VIS transformer}. 
	In the feature extraction stage, given a query video and a handful of (self-shot) support videos, the backbone extracts features of individual image frames, then the image features are concatenated in the frame order to form clip-level features for the query and support videos. In the common transformer, the encoder models the pixel-level similarity for the query and support features respectively, the fuser leverages the similarity between the query feature and the support feature, and the decoder learns the similarity between instances along the time dimension. In the prediction stage, the instance sequences are inferred in the query video 
	} 
	\label{fig:architecture}
\end{figure}

\paragraph{Stage 1: Spatio-temporal transformer encoder.} 
We first feed the extracted video features into the transformer encoder structure and flatten the spatial and temporal dimensions of $f_q$ and $f_s$ in 2D feature maps of size $d {\times} ( T {\cdot} W {\cdot} H )$ and $ d {\times}( T' {\cdot} W {\cdot} H )$. 
Since the image-encoder based backbone is permutation-invariant, we append spatio-temporal positional encodings~\cite{bello2019attention,parmar2018image} to the inputs as the instance segmentation task requires precise spatial and temporal information.
Specially, for all spatio-temporal coordinates of each dimension, we independently use $d/3$ sine and cosine functions with different frequencies. 
We then concatenate them to get the final $d$ channel encoding. The spatio-temporal positional encodings are added to both the query feature and support feature in each encoder layer. The output of this transformer encoder structure for the query branch is $E_q\in \mathbb{R}^{d {\times} (T {\cdot} W {\cdot} H)}$, and the output for the support branch is $E_s\in \mathbb{R}^{d {\times} (T' {\cdot} W {\cdot} H)}$. The encoder weights are also shared.

\paragraph{Stage 2: Query-support fuser.}
Given encoded query and support videos, we seek to discover similarities in space and time by integrating the support branch into the query branch, by utilizing the attention mechanism.
Let $\mathbb{MA}$ denote the multi-headed attention with linear projection function $Q(\cdot)$, $K(\cdot)$, $V(\cdot)$ as described in~\cite{vaswani2017attention}.
We first cross-enhance the fuser inputs $E_q$ and $E_s$ through multi-head attention, as shown in Figure~\ref{appa:fig:module} in the Appendix:
$
f_{q \gets s}=\text{LN}(E_q + \mathbb{MA}(Q(E_q),K(E_s),V(E_s))
$, and similarly for $f_{s \gets q}$.
%
Here, LN denotes the layer normalization operation~\cite{ba2016layer}. 
Next, the support branch is fused into the query branch to get the fused feature $\widetilde{F}$: 
\begin{equation}
\widetilde{F}=\text{LN}(f_{q \gets s} + \mathbb{MA}(Q(f_{q \gets s}),K(f_{s \gets q}),V(f_{s \gets q})).
\label{equ:F_tilde}
\end{equation}
In addition, a feed-forward network (2-layer MLP) is applied to $\widetilde{F}$ in a residual fashion for increased modelling ability, yielding output of the fuser:
$
\mathcal{F}=\text{LN}(\widetilde{F}+\text{FFN}(\widetilde{F})).
$

\paragraph{Stage 3: Decoding and predicting.} The spatio-temporal decoder aims to decode the most discriminative pixel features that can represent the instances of each frame. We introduce a fixed number of input embeddings to represent the instance features across time and space, which we call \textit{instance  sequences}. 
Assuming that the model decodes $n$ instances per frame, the number of instances for the $T$ frames in the query video is $N {=} n {\cdot} T$. 
The \textit{instance  sequences} are learned by the spatio-temporal decoder 
and take the output of the query-support fuser $\mathcal{F}$ and \textit{instance  sequences} as input, to outputs $N$ instance features, denoted as $\mathcal{D}$, as shown in Figure~\ref{fig:architecture}.
%
%
Finally, the instance segmenter predicts the mask sequence for each instance. 
For each frame in the query video, we feed the instance features $\mathcal{D}$ and the fused feature $\mathcal{F}$ into an attention module to obtain the attention maps. 
The attention maps are then concatenated with the encoded query feature $E_q$ and the fused feature $\mathcal{F}$, followed by a deformable convolution layer~\cite{dai2017deformable}. 
In this way, we obtain the mask features for each instance of the different frames in the query video. We denote the mask feature for instance $i$ of frame $t$ is $g_{i,t} \in \mathbb{R}^{1 \times a \times W_0 \times H_0}$, where $a$ is the channel number, $W_0$ and $H_0$ are the feature width and height. 
Finally, the instance segmenter uses the accumulated features to output the mask sequence for each instance (see Appendix for details).
and outputs the mask sequence $m_i \in \mathbb{R}^{1 \times 1 \times T \times W_0 \times H_0}$ for the instance $i$ directly. 

\paragraph{Training loss.}
To score predicted instances with respect to the ground truth, we introduce an optimal bipartite matching between predicted and ground truth instances, in the spirit of~\cite{carion2020detr,wang2021end} (see Appendix for details).
%
Given the optimal assignment, the next step is to compute the training loss $\mathcal{L}_{\mathrm{train}}$, which is a linear combination of a negative log-likelihood for \textit{foreground/background} prediction, a box loss and mask loss for the instance sequences:
\begin{equation}
\begin{split}
\mathcal{L}_{\mathrm{train}} & (y, \hat{y}) = \sum_{i=1}^{n}[(-\log\hat{p}_{\hat{\sigma}(i)}(c_i))+\\
&\mathcal{L}_{\mathrm{box}}(b_i,\hat{b}_{\hat{\sigma}(i)})+\lambda_{\mathrm{mask}} \cdot \mathcal{L}_{\mathrm{mask}}(m_i,\hat{m}_{\hat{\sigma}(i)})],
\label{equ:loss}
\end{split}
\end{equation}
here $c_i {=} \textit{foreground}$, and $\hat{\sigma}$ is the optimal assignment, $\hat{p}_{\sigma(i)}(c_i)$ denotes the probability of $c_i$ with index $\sigma(i)$, $b_i$ denotes the ground truth box sequences.
This training loss is used to train the whole video instance segmentation framework end-to-end. 
For the bounding box loss we employ the generalize IoU loss as prescribed in~\cite{rezatofighi2019generalized}, while we use a linear combination of the dice loss~\cite{milletari2016v} and focal loss~\cite{lin2017focal} for the mask loss. 
The full loss equations are provided in the Appendix.
As a result we obtain an end-to-end framework that is guided by support videos and able to segment instances in a query video. 


\textbf{Implementation details.} As the largest video length in Self-VIS is $32$, we take $32$ as the query video clip length $T$. The support video clip length is set to $24$. If the original video is too short or too long, we pad it with the last frame or cut it at a random position. All videos are resized to a $320{\times}280$ resolution before they are fed into the backbone. The model predicts 10 instances for each query frame, thus the total instance number is $320$. In the common transformer structure, we use $6$ encoder, $3$ fuser, $6$ decoder layers of width $288$ with $8$ attention heads. The model is trained with AdamW~\cite{loshchilov2017decoupled}, setting the initial common transformer's learning rate to $10^{-4}$, the backbone's learning rate to $10^{-5}$. 
The model is trained for $20$ epochs, with the learning rate decay by $10$ at $14$ epochs. We initialize our backbone with the weights pretrained on the COCO dataset~\cite{lin2014microsoft}. Further details are provided in the Appendix.

The evaluation metric is average precision, with the video Intersection over Union (IoU) of the mask sequences as threshold. The IoU threshold is set to $0.5$ unless specified otherwise. All code and scripts for reproducing experiments will be released.

\section{Results}

\textbf{Self-shot evaluation.}
We first evaluate the potential of self-shot learning on the introduced video instance segmentation benchmarks. We use the introduced transformer and compare to both oracle upper bounds and self-supervised baselines for obtaining support videos as input to the transformer.

In Table~\ref{tab:self_support}, we compare a broad range of feature spaces for providing relevant support videos given a query video. 
We first find that utilizing random videos from the unlabelled dataset as support already provides a non-trivial instance segmentation performance (row \textit{a}), but still with a considerable gap to the oracle baseline where each support video is manually curated to have matching instance classes with the query video. Self-shot learning brings large benefits over randomly picking support videos (rows \textit{(c)-(f)}).
We first establish a baseline of using ImageNet-pretrained features for obtaining relevant support samples in row \textit{(c)}, as well as finetuning these features using a non-parametric instance retrieval loss~\cite{wu2018unsupervised} using MoCo~\cite{he2020momentum} in row \textit{(d)}, which adds around 2.5-3.5\% in performance.
Next, we utilize a more sophisticated MIL-NCE loss~\cite{miech2020end} and a rank-based retrieval loss~\cite{varamesh2020self} to learn \textit{video-clip} embeddings (see Appendix for details).
With this we find that row \textit{(f)} in Table~\ref{tab:self_support} achieves strong gains of 6-10\% in absolute performance compared to the random baseline and more than 3-7\% compared to frozen ImageNet features and use this for subsequent experiments.
Besides the finding that all features obtained in a self-supervised fashion improve over the supervised frame-based features ones we also establish that the self-supervised task does matter too, as we find the ranking loss well-suited for retrieving relevant support videos. 
We conclude that the strongest self-support can almost close the gap with the oracle-support baseline, even though self-shot support videos are not guaranteed to have matching classes.



\begin{table}[t]
\centering
	\captionof{table}{\textbf{Self-shot evaluation.} The unsupervised support videos come from the Youtube-8M Segments dataset and video instance segmentation performance is evaluated on the test set of Self-VIS. For comparability, we include baselines using random videos or fully supervised oracle videos as support. Inference-time support-increase is evaluated in the $5+(n)$ columns where extra $n$ support videos are used at inference. The strongest self-support is competitive with a 1-shot oracle-support}
   	\label{tab:self_support}
   	\setlength{\tabcolsep}{3pt}
		\begin{tabular}{lccccc}
			\toprule
			 & \multicolumn{5}{c}{\textbf{Support}} \\
			\cmidrule(lr){2-6}
			  & 1 & 5 & 5+(1) & 5+(3) & 5+(5)\\
			\midrule
		\rowcolor{mygray}
		\textbf{Retrieval based on }& & & & & \\
		\textit{(a)}\quad Random videos     & 44.3 & 44.9 & 44.8 & 45.1 & 45.1\\
		\textit{(b)}\quad Oracle/labels     & 53.2 & 56.6 & 56.1 & 57.6 & 57.8\\
		\midrule
		\rowcolor{mygray}
		\textbf{Self-shot variants }& & & & & \\
		\textit{(c)}\quad  Inception fixed  & 46.9 & 48.3 & 48.5 & 48.2 & 48.4\\
		\textit{(d)}\quad  Inception MoCo   & 49.4 & 51.6 & 51.9 & 51.7 & 51.9\\
		\textit{(e)}\quad  Video MIL-NCE    & 50.1 & 52.5 & 52.6 & 52.9 & 53.3\\
		\textit{(f)}\quad  Video Rank       & 51.4 & 54.3 & 54.6 & 55.2 & 55.4\\
		\bottomrule
		\end{tabular}%
\end{table}
    
\textbf{Self-shot versus zero-shot learning.}
To further quantify the effectiveness of self-shot learning for finding support videos, we compare to zero-shot learning on two different tasks: video instance segmentation and temporal action localization. 
For video instance segmentation, all experiments run on the Self-VIS dataset and as a zero-shot baseline we utilize the method from Dave~\etal~\cite{dave2019towards} that can segment moving objects in videos, even the ones unseen in training. For temporal action localization, we follow the setup of Yang~\etal~\cite{yang2020localizing}, which also provides the temporal action localization pipeline and the reorganized dataset derived from Thumos14. Zhang~\etal~\cite{zhang2020zstad} provide the zero-shot method for temporal action localization. They can localize the unseen activities in a long untrimmed video based on the label embeddings, which means class labels are needed during inference. In Table~\ref{temporal_action}, we report the video instance segmentation and temporal action localization results. We find that on both settings our self-shot approach improves over the zero-shot alternatives, indicating that automatically obtaining support videos in an unsupervised manner and using them for their respective video task obtains favorable results over a semantic transfer of information from seen to unseen visual classes. In the Appendix, we also provide results where we use zero-shot learning to help find support videos, which is also not as effective as self-shot learning.

\begin{table}[t] 
\footnotesize
        \centering
        \caption{\textbf{Self-shot versus zero-shot learning} for video instance segmentation and temporal action localization. 
        For Temporal action localization, we follow the setup of Yang~\etal~\cite{yang2020localizing} on the Thumos14 dataset. The metric is video-mAP with an overlap threshold of 0.5. We find that our self-shot perspective is better suited for segmentation and localization in videos than zero-shot baselines}
        \label{temporal_action}
        \begin{tabular}{l c ccc c ccc}
        \toprule
        &&      \multicolumn{3}{c}{\textbf{Self-VIS}} && \multicolumn{3}{c}{\textbf{Thumos14}} \\
        \cmidrule{3-5} \cmidrule{7-9}
        && 0& 1 & 5 && 0& 1 & 5 \\
        \midrule
        Zero-shot learning~\cite{dave2019towards} &&47.4&-&-&&-&-&-\\
        Zero-shot learning~\cite{zhang2020zstad}  &&-&-&-& &43.4&-&- \\
        Self-shot learning                       &&-& 51.4  & 54.3  &&-& 45.8 & 47.3 \\
        Self-shot learning k+(5)                 &&-& 54.6  & 55.4  &&-& 47.7 & 48.0 \\
        \bottomrule
        \end{tabular}
\end{table}


\textbf{Transformer ablation.}
\begin{table}[t!]
	\centering
	\caption{\textbf{VIS transformer ablation under self-shot setting.} The decoder achieves competitive performance by itself. It improves when the encoder processes the videos. A considerable performance gain happens when the fuser passes messages from the support branch to the query branch. Performance is best with all three modules
}
	\label{tab:self_ablation_module}
	\resizebox{0.8\columnwidth}{!}{%
	\begin{tabular}{ccccccc}
	\toprule
	\multirow{2}*{\textbf{Encoder}} & \multirow{2}*{\textbf{Fuser}} & \multirow{2}*{\textbf{Decoder}} & \multicolumn{2}{c}{\textbf{Self-VIS}} & \multicolumn{2}{c}{\textbf{Self-OVIS}}\\
		\cmidrule(lr){4-5} \cmidrule(lr){6-7}
	  &  &  & 1 self-shot & 5 self-shots & 1 self-shot & 5 self-shots\\
	\midrule
	& & $\checkmark$ & 40.7  & 40.8 & 13.7 & 14.4\\
    $\checkmark$ & & $\checkmark$ & 42.4 & 42.2 & 14.3 & 15.7\\
    & $\checkmark$ & $\checkmark$ & 49.1 & 51.5 & 19.6 & 21.0\\
	$\checkmark$ & $\checkmark$ & $\checkmark$ & \textbf{51.4} & \textbf{54.3}& \textbf{20.6} & \textbf{23.7}\\
	\bottomrule
    \end{tabular}%
	}
\end{table}
In Table~\ref{tab:self_ablation_module}, we show the effect of the three components in our video instance segmentation transformer when using both one and five self-shot support videos. We find that all three components matter for maximizing segmentation performance. Especially the introduced fuser module is important and adds $8.4$\% 1-shot performance compared to the baseline, while the encoder adds $1.7\%$. When combined, the encoder yields an additional $2.3\%$ gain on top of the fuser-only baseline, showing that having sufficient encoding capacity before fusing leads to better performance. In the appendix we also establish that the proposed baseline is competitive both against previous works adapted for the novel video instance segmentation setting, as well as against established methods for an image segmentation setting. Overall, we conclude that the proposed transformer is effective for video instance segmentation using a few support videos.


\textbf{Video difficulty ablation.}
\begin{table}[t]
\setlength{\tabcolsep}{2pt}
\footnotesize
\centering
\caption{\textbf{Video difficulty ablation.} 
\textbf{(a)}: Performance when varying the number of instances on Self-VIS. High scores can be obtained when the common instances are not too many (no more than 3), segmentation of more than 4 instances in a query video remains challenging.
\textbf{(b)}: Performance when varying the number of classes on Self-OVIS. When the common instances come from multiple classes, segmentation becomes harder}
\label{tab:ablation_query}
\subfloat[\label{tab:instance_number}Instance number in query video]{
\begin{tabular}{cccccc}
\toprule
			&\multicolumn{5}{c}{\textbf{Instance number}}\\
			\cmidrule(lr){2-6}
			 & 1& 2 & 3 & 4 & $\geq$5 \\
			\midrule
		1 self-shot & 52.1 & 51.7 & 49.5 & 44.7 & 41.6  \\
		5 self-shots & 54.8 & 54.6 & 52.9 & 50.2 & 45.5 \\
		\bottomrule
\end{tabular}
}%
\hspace{5mm}
\subfloat[\label{tab:class_number}Class number in query video]{
\begin{tabular}{ccccc}
\toprule
			&\multicolumn{4}{c}{\textbf{Class number}}\\
			\cmidrule(lr){2-5}
			 & 1& 2 & 3 & $\geq$4\\
			\midrule
		1 self-shot & 23.4 & 20.7 & 17.1 & 8.6  \\
		5 self-shots & 26.2 & 23.3 & 19.4 & 9.4\\
		\bottomrule
\end{tabular}
}
\end{table}
Next, we ablate the effect of the number of instances and number of classes in the query video on the segmentation performance in Table~\ref{tab:ablation_query}. 
As each video in the Self-VIS dataset contains instances from just a single class, we use this to analyze the stability of our model with regard to the number of instances.
The result is shown in Table~\ref{tab:instance_number}, and we find performance only mildly declines for a moderate increase, up to 3, in number of instances in the query video. 
Next, we use the more difficult Self-OVIS dataset to study the robustness against more diverse videos that contain more instances from multiple classes. 
As shown in Table~\ref{tab:class_number}, the method still works well with more than one instance per video, though performance naturally declines with this added difficulty for the task.

\begin{figure}[t]
    \centering
    \includegraphics[width=0.65\columnwidth]{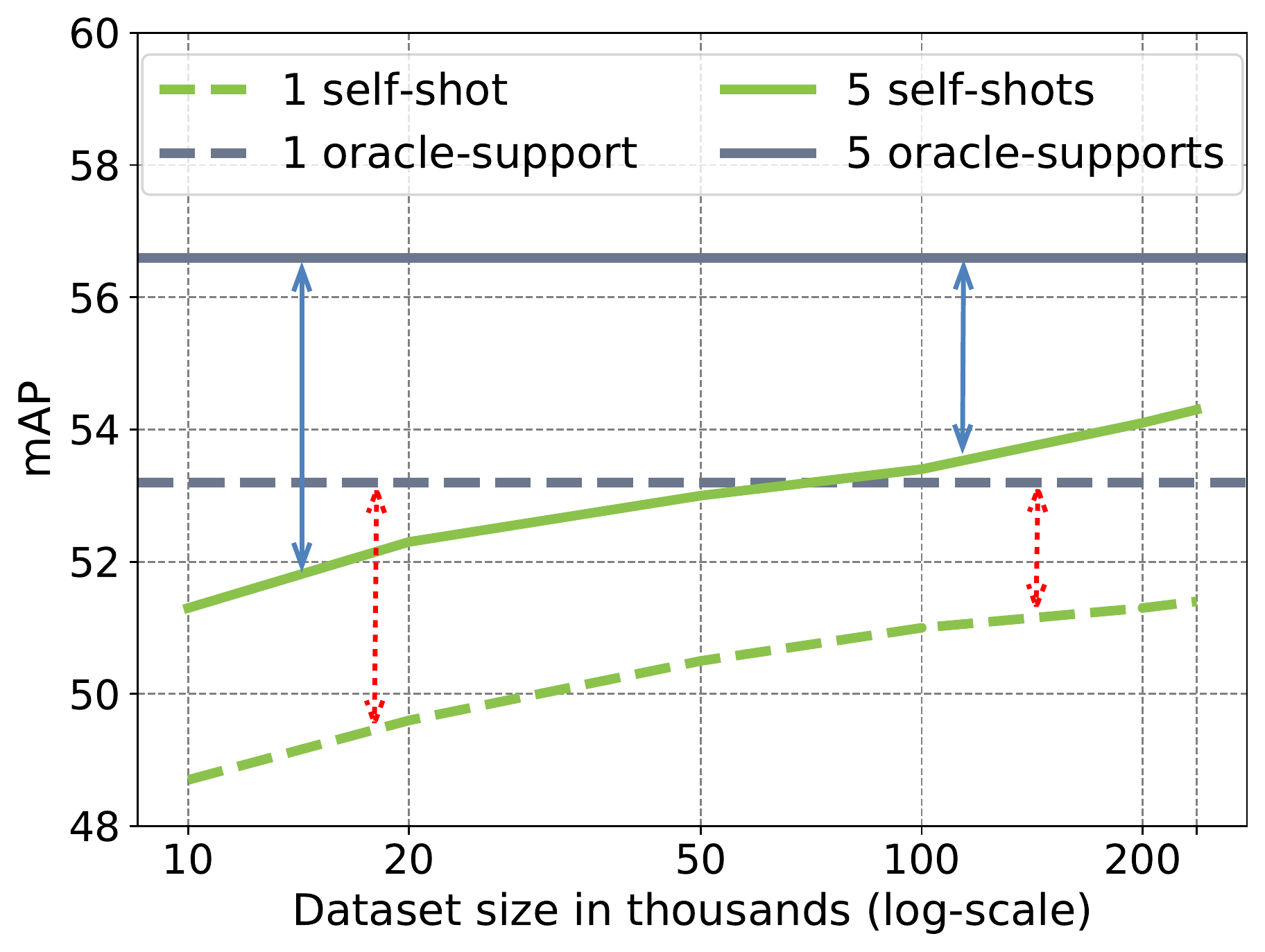}
	\caption{\textbf{Scalability of self-shot learning.}
Performance scales positively with the increase in unlabelled videos available and 5 self-supports outperforms using 1 oracle-support for $>$75K videos
}
\vspace{-0.5em}
	\label{fig:dataset_size}
\end{figure}

\begin{figure}[t]
	\centering
	\includegraphics[width=0.88\columnwidth]{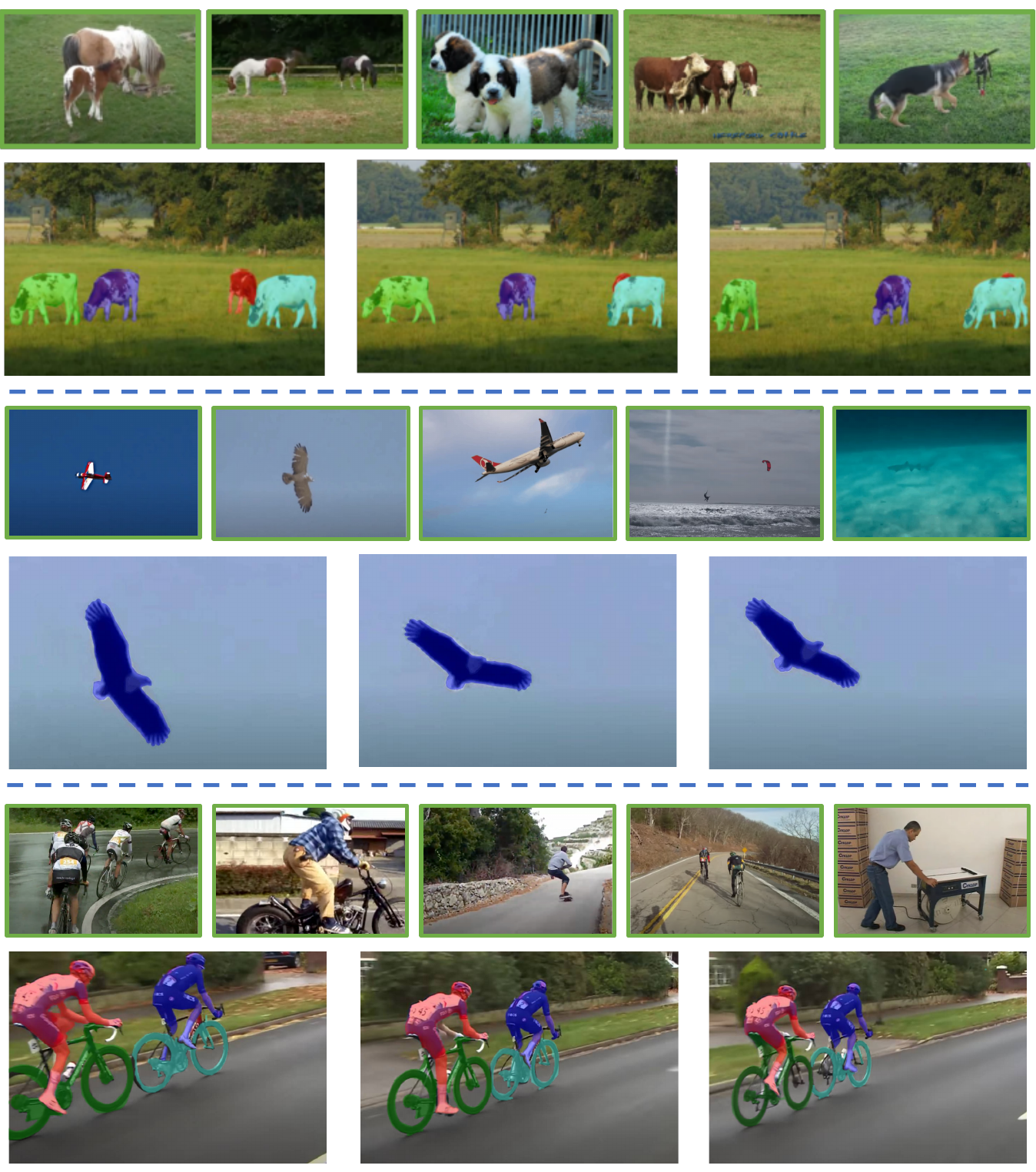}
	\caption{\textbf{Qualitative examples.} Three examples of self-shot videos (top) and the resulting instance segmented query video (bottom)}
	\label{fig:examples}
\end{figure}

\textbf{Scaling unsupervised support.}
The quality of the self-shot learning is bounded by the quality of the videos in the unlabelled video collection. In Figure~\ref{fig:dataset_size} we show how self-shot video instance segmentation scales with unlabelled dataset size. 
We find that at around 75K videos, our self-shot approach with 5 videos already \textit{outperforms the 1-shot oracle-support} baseline.
While we cannot perform experiments using even larger dataset sizes, we can see that even at 237K videos, the performance is still rising steeply on a typical log-datasize scale. 
Based on this result, oracle-support from a labelled dataset of limited size might not even present a top-line for the same number of support videos and as we have shown in Table~\ref{tab:self_support} can be further boosted with inference-time increase in number of supports. 
Thus, this shows that using larger unlabelled video datasets and conducting retrieval presents an effective method for scalable video instance segmentation. In Figure~\ref{fig:examples} we provide qualitative examples of self-shot learning.

\textbf{Semi-shot learning.}
In the final experiment, we show results for combining self-shot and oracle support videos. In this setting, we are given a few oracle support videos and use these to help find more support videos in a self-shot manner, arriving at a semi-shot alternative. In Table~\ref{tab:semi-shot}, we show using a single oracle video and $4$ self-shot videos boosts performance by $1.9\%$ and achieves $55.1\%$ mAP, thus almost reaching the performance when using 5 oracle-support videos, all without any further annotation requirements.

\begin{table}[t]
\centering
\caption{\textbf{Semi-shot learning.} We show that the performance increases
	when adding additional self-shot videos to the oracle-support videos, arriving at a semi-shot alternative. We find that such a hybrid setting can quickly close the gap to few-shot learning with ground truth support videos, further highlighting the potential of self-shot learning for video instance segmentation}
	\label{tab:semi-shot}
	\setlength{\tabcolsep}{4pt}
		\begin{tabular}{lccccccc}
			\toprule
			& &\multicolumn{6}{c}{\textbf{\# Oracle-support}}\\
			\cmidrule(lr){3-8}
			 & & 0& 1 &2 & 3& 4&5\\
			\midrule
		\multirow{6}*{\rotatebox{90}{\textbf{\# Self-shot}}}		&0&     & 53.2 & 53.9& 55.0& 55.7& 56.6\\
		&1& 51.4& 53.6 & 54.5& 55.1 & 56.0&\\
		&2& 52.3& 54.2 & 54.8& 55.9 &&\\
		&3& 52.8& 54.7 & 55.2 &&&\\
		&4& 53.6& 55.1 &&&\\
		&5& 54.3&    &&&  \\    
		\bottomrule
		\end{tabular}%
\end{table}

\section{Discussion}
\paragraph{Limitations.}
While we have proposed and explored the task of self-shot video instance segmentation, we have done so in a sequential fashion: The support set generation is detached from the segmentation pipeline. While this allows for better analysis of the method, performance can likely be further improved by training in an end-to-end fashion. We are also limited by the unlabelled video dataset size because of our computational resources. 
As shown in Figure~\ref{fig:dataset_size}, larger dataset sizes will likely further highlight the benefit of utilizing self-shot learning.


\paragraph{Conclusions.}
This paper proposed the task of self-shot learning.
We have analysed and proposed this in the most annotation intense setting, that of video instance segmentation where existence of oracle support can be considered unrealistic. 
For this we develop a novel transformer based instance segmentation baseline and outline how to obtain support videos automatically from an unlabelled pool through self-supervision. Experiments show that our approach achieves strong performance, can already outperform oracle support in some settings, is scalable, and can be combined in a semi-supervised setting.

\clearpage

\bibliography{7_egbib}
\clearpage
\clearpage
\appendix

\section{Further Details}

\subsection{Datasets}
\textbf{Dataset statistics.} We report the statistics of the self-shot video instance segmentation datasets in Table~\ref{appx:tab:dataset}. 
We list the class names for the train, validation and test sets in Table~\ref{appx:tab:subset}.

\begin{table}[t] 
        \centering
        \caption{\textbf{Overview of the common video instance segmentation datasets.} Self-VIS contains a few instances of one single class per video, while Self-OVIS contains videos with more instances of multiple classes}
        \label{appx:tab:dataset}
         \resizebox{0.7\columnwidth}{!}{
        \begin{tabular}{lrr}
        \toprule
         & \textbf{Self-VIS} & \textbf{Self-OVIS} \\
        \midrule
        \textbf{Video statistics} & & \\
        \quad mean length (frame) & 30.2 & 69.4 \\
        \quad mean class number & 1 & 1.6 \\
        \quad mean instance number & 1.7 & 5.9 \\
        \quad number of train videos & 1{,}651 & 449  \\
        \quad number of val+test videos & 472 & 158  \\
        \midrule
        \textbf{Class statistics} & & \\
        \quad number of train classes & 30 & 17  \\
        \quad number of val+test classes & 10 & 8  \\
        \bottomrule
        \end{tabular}
        }
\end{table}

\begin{table*}
        \centering
        \caption{\textbf{Subset labels on self-shot video instance segmentation datasets.} Performance numbers in the paper refer to models evaluated on the test splits. Note that we do not use the labels in our experiments}
        \label{appx:tab:subset}
         \resizebox{1\columnwidth}{!}{
        \begin{tabular}{p{.18\textwidth}p{.55\textwidth}p{.02\textwidth}p{.35\textwidth}}
        \toprule
          & \textbf{Self-VIS} & & \textbf{Self-OVIS} \\
        \midrule
        \textit{\textbf{Training}} & {TennisRacket, Snowboard, Surfboard, EarlessSeal, Horse, Boat, Mouse, Tiger, Frog, Elephant, Truck, Owl, Airplane, Zebra, Train, Deer, Fish, Leopard, Turtle, Fox, Duck, Snake, Skateboard, Rabbit, Cat, Sedan, Parrot, GiantPanda, Ape, Person } & & {Fish, Vehical, Sheep, Zebra, Bird, Poultry, Elephant, Motorcycle, Dog, Monkey, Boat, Turtle, Cow, Parrot, Giraffe, Tiger, GiantPanda} \\
        \midrule
        \textit{\textbf{Validation and testing}} & {Eagle, Cow, Motorbike, Bear, Dog, Shark, Giraffe, Monkey, Lizard, Hand} & & {Person, Horse, Bicycle, Rabbit, Cat, Airplane, Bear, Lizard}    \\ 
        \bottomrule
        \end{tabular}
        }
\end{table*}

\subsection{Self-shot construction}
\textbf{Ranking loss.} $R_a(b,c)$ in Equation~\textcolor{red}{2} in the paper is a differentiable function to approximately rank video $b$ among all videos in the set $\{c\}$ with respect to the query video $a$. 
It was first introduced for images in~\cite{varamesh2020self}.
\begin{equation}
R_a(b,c) = 1+\sum_{\hat{c} \in \{c\},\hat{c} \neq b}\frac{1}{1+\exp(-d_{a\hat{c}b}/2)},
\label{equ:rank}
\end{equation}
\begin{equation}
d_{a\hat{c}b}=\frac{a\cdot\hat{c}}{\Arrowvert a \Arrowvert\cdot\Arrowvert\hat{c}\Arrowvert}-\frac{a \cdot b}{\Arrowvert a \Arrowvert\cdot\Arrowvert b \Arrowvert}.
\label{equ:distance}
\end{equation}
With this approximation we are able to design the differentiable ranking loss $\mathcal{L}_{Rank}$ of Equation~\textcolor{red}{2}.

\subsection{Self-shot VIS transformer}
\textbf{Feature extraction.} We adopt a modified ResNet-50~\cite{he2016deep} with a bigger receptive field for feature extraction. Specifically, we remove the last stage of the ResNet-50 and take the outputs of the fourth stage as final outputs. We further modify the $3{\times}3$ convolution in the fourth stage to a dilated convolution with a stride of $2$ instead of $1$ to increase the receptive field. Finally, a learned $1 {\times} 1$ convolution is applied to reduce the channel dimension. During training, the modified layers are initialized with Xavier~\cite{glorot2010understanding} init, all other backbone parameters are initialized with a COCO-pretrained ResNet-50.

\begin{figure}[t!]
	\centering
	\includegraphics[width=0.6\columnwidth]{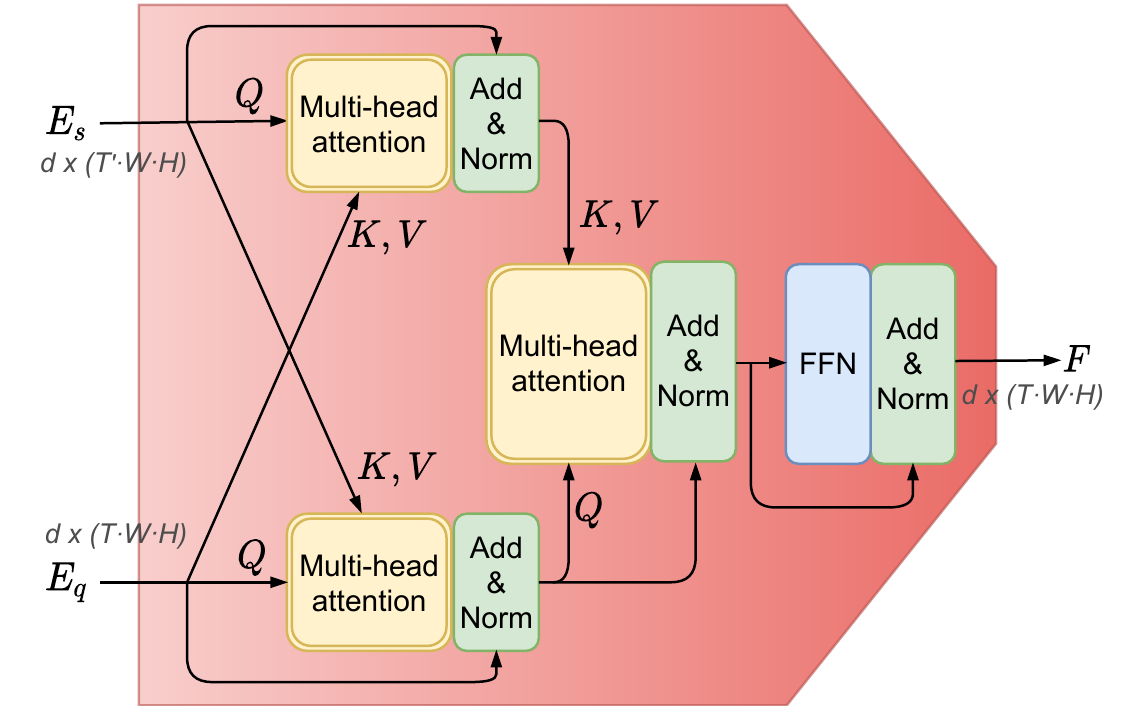}
	\caption{\textbf{Structure of query-support fuser.} First, the query and support representations are enhanced by each other simultaneously through message passing. Then, the support branch is fused into the query branch by the multi-head cross-attention mechanism. Finally, a feed-forward neural network (FFN) module in the form of a residual is performed to augment the fitting ability. \textbf{Q, K, V} denote different roles in the multi-head attention mechanism}
	\label{appa:fig:module}
\end{figure}

\begin{figure}[tb!]
	\centering
	\includegraphics[width=1.0\columnwidth]{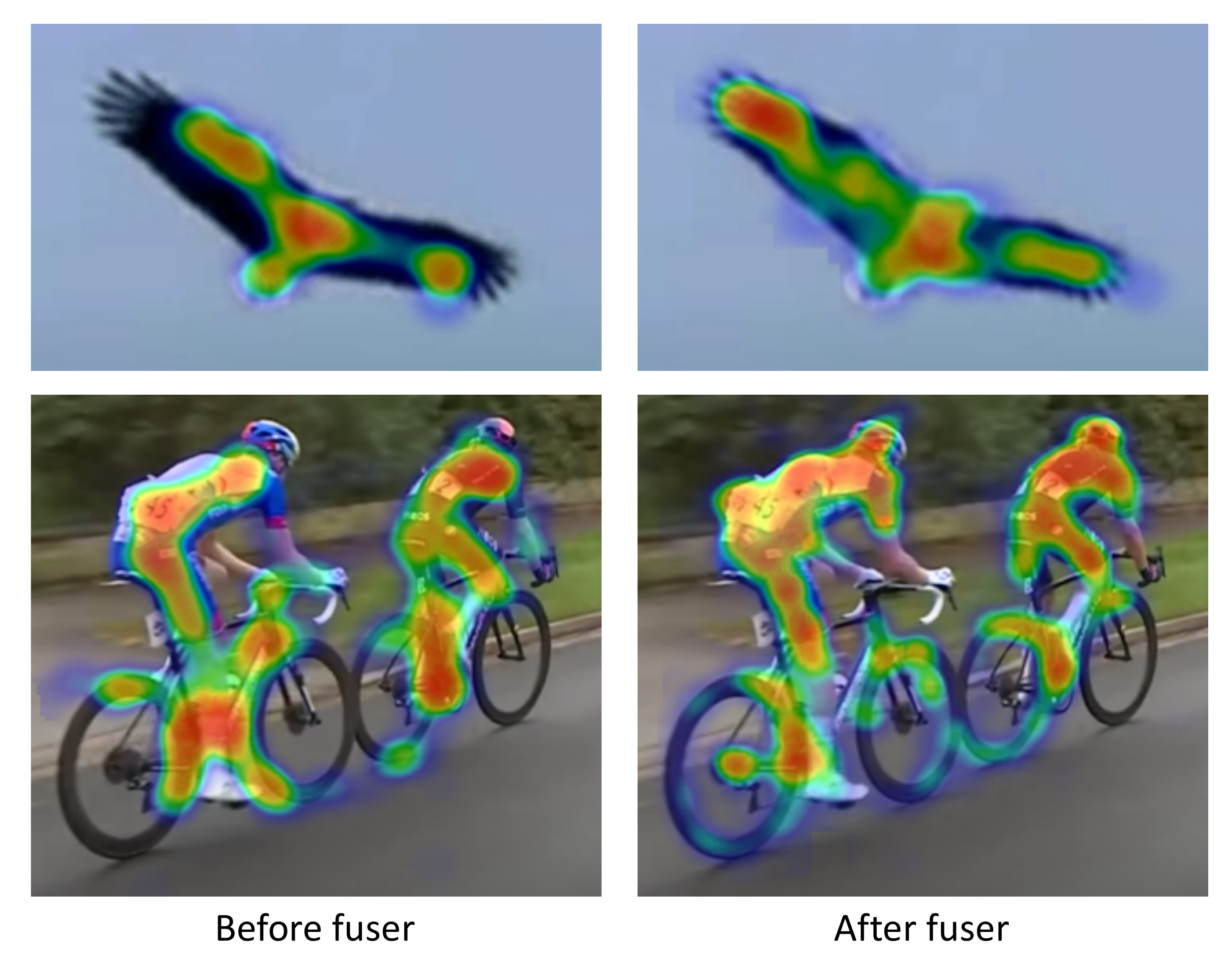}
	\caption{\textbf{Attention heatmap visualization} before and after the fuser module. For both examples, the left and right show the attention heatmaps before and after the fuser module. The heatmap results show the fuser module can better highlight the individual instances in the query video with the aid of supports.}
	\label{fig:heat}
\end{figure}

\textbf{Query-support fuser.} The query-support fuser can effectively leverage the similarity between the query and support videos, the detailed structure of which is shown in Figure~\ref{appa:fig:module}. 
First, the query and support representations are enhanced by each other simultaneously through multi-head attention. 
Then, the support branch is fused into the query branch by the multi-head cross-attention mechanism. Finally, a feed-forward neural network (FFN) module in the form of a residual is performed to augment the fitting ability. To better understand the effectiveness of the fuser module, we visualize the attention heatmaps before and after the fuser module in Figure~\ref{fig:heat}. The figure shows that the proposed fuser module is able to better highlight the instances of interest in the query video with the aid of supports.

\textbf{Instance segmenter.} The instance segmenter predicts the mask sequence for each instance. To predict mask sequences that are informed across time, we predict the masks based on the accumulation of mask features per instance. For each frame in the query video, we feed the instance features $\mathcal{D}$ and the fused feature $\mathcal{F}$ into an attention module to obtain the attention maps. 
Then the attention maps are concatenated with the encoded query feature $E_q$ and the fused feature $\mathcal{F}$, followed by a deformable convolution layer~\cite{dai2017deformable}. 
In this way, we obtain the mask features for each instance of different frames in the query video. 
Assume that the mask feature for instance $i$ of frame $t$ is $g_{i,t} \in \mathbb{R}^{1 \times a \times W_0 \times H_0}$, where $a$ is the channel number, $W_0$ and $H_0$ are the feature width and height. 
Then we concatenate the mask features of $T$ frames in the query video to form the $G_i \in \mathbb{R}^{1 \times a \times T \times W_0 \times H_0}$. 
The instance segmenter takes the instance mask features $G_i$ as input, and outputs the mask sequence $m_i \in \mathbb{R}^{1 \times 1 \times T \times W_0 \times H_0}$ for the instance $i$ directly. 
The instance segmenter consists of three 3D convolutional layers and group normalization layers~\cite{wu2018group} with ReLU activation function, yielding $T$ mask predictions for each instance $i$.

\textbf{Optimal bipartite matching.} Let $\hat{y}{=}\{ \hat{y} \}_{i=1}^n$ denote the predicted instance sequences, and $y$ the ground truth set of instance sequences. 
Assuming $n$ is larger than the number of instances in the query video, we consider $y$ also a set of size $n$ by padding with $\emptyset$ (background). To find a bipartite matching between the two sets we search for a permutation of $n$ elements $\sigma $ with the lowest cost.
As computing the instance mask sequence similarity directly is computationally intensive, we replace the mask sequence with the box sequence to perform the matching. Following the same practice of~\cite{carion2020detr,yang2021few}, we get the normalized center coordinates, height and width of the boxes and the binary \textit{foreground/background} labels. The matching function is  defined as~\cite{carion2020detr}:
\begin{equation}
\mathcal{L}_{\mathrm{match}}(y_i,\hat{y}_{\sigma(i)}) = -\hat{p}_{\sigma(i)}(c_i)+\mathcal{L}_{\mathrm{box}}(b_i,\hat{b}_{\sigma(i)}),
\label{equ:match}
\end{equation}
where $c_i {=} \textit{foreground}$, $\hat{p}_{\sigma(i)}(c_i)$ denotes the probability of $c_i$ with index $\sigma(i)$, $b_i$ denotes the ground truth box sequences, and $\hat{b}_{\sigma(i)}$ denotes the predicted box sequences with index $\sigma(i)$. 
Based on the above criterion, a one-to-one matching of the sequences is found by the Hungarian algorithm~\cite{kuhn1955hungarian}, following prior work~\cite{stewart2016end,carion2020detr}.

\textbf{Components of training loss.} For the bounding box loss $\mathcal{L}_{\mathrm{box}}$, we use a linear combination of the sequence level $L1$ loss and the generalized IoU loss~\cite{rezatofighi2019generalized}:

\begin{equation}
\begin{split}
\mathcal{L}_{\mathrm{box}}(b_i,\hat{b}_{\hat{\sigma}(i)}) &= \frac{1}{T}\sum_{t=1}^T[\lambda_{iou} \cdot \mathcal{L}_{\mathrm{iou}}(b_{i,t},\hat{b}_{\hat{\sigma}(i),t})\\
&+\lambda_{L1}\Arrowvert b_{i,t}-\hat{b}_{\hat{\sigma}(i),t} \Arrowvert_1],
\label{equ:box_loss}
\end{split}
\end{equation}

The mask loss is defined as a combination of the dice loss~\cite{milletari2016v} and focal loss~\cite{lin2017focal}:

\begin{equation}
\begin{split}
\mathcal{L}_{\mathrm{mask}}(m_i,\hat{m}_{\hat{\sigma}(i)}) &= \frac{1}{T}\sum_{t=1}^T[ \mathcal{L}_{\mathrm{dice}}(m_{i,t},\hat{m}_{\hat{\sigma}(i),t})\\
&+\mathcal{L}_{\mathrm{focal}}(m_{i,t},\hat{m}_{\hat{\sigma}(i),t})],
\label{equ:mask_loss}
\end{split}
\end{equation}

\section{Further results}
\subsection{Few-shot experiments with oracle supports}

\textbf{Ablation of transformer depth.} 
We ablate the effect of the layer depth of the encoder, fuser and decoder modules in Table~\ref{tab:ablation_size} under the one- and five-shot setting on  Self-VIS. We find the increase in performance is steady across all three modules when adding further capacity and use the setting of 6 encoder layers, 3 fuser layers and 6 decoder layers for all other experiments.

\begin{table*}[t!]
\centering
\caption{\textbf{Ablation of transformer depth} of the encoder, fuser and decoder on Self-OVIS under the one- and five- shot setting. The default layer number is 6,3,6 for encoder, fuser and decoder (highlighted in gray). When we ablate the layers of one module, the other two modules have default layers. Performance improves gradually with more layers in the modules}
\label{tab:ablation_size}
\resizebox{1\columnwidth}{!}{%
\subfloat[\label{tab:encoder_size}Encoder size]{
\begin{tabular}{ccccc}
\toprule
			&\multicolumn{4}{c}{\textbf{Encoder layers}}\\
			\cmidrule(lr){2-5}
			 & 2& 4 & 6 & 8  \\
			\midrule
		One-shot & 51.7 & 52.4 & \colorbox{lightgray}{53.2} & 53.2   \\
		Five-shot & 55.1 & 56.1 & \colorbox{lightgray}{56.6} & 56.4  \\
		\bottomrule
\end{tabular}
}%
\hspace{3mm}
\subfloat[\label{tab:fuser_size}Fuser size]{
\begin{tabular}{cccccc}
\toprule
			&\multicolumn{5}{c}{\textbf{Fuser layers}}\\
			\cmidrule(lr){2-6}
			 & 1 & 2 & 3 & 4 & 5 \\
			\midrule
		One-shot & 51.3 & 52.5 & \colorbox{lightgray}{53.2} & 53.4 & 53.5 \\
		Five-shot & 54.0 & 55.8 & \colorbox{lightgray}{56.6} & 56.9 & 57.1 \\
		\bottomrule
\end{tabular}
}
\hspace{3mm}
\subfloat[\label{tab:decoder_size}Decoder size]{
\begin{tabular}{ccccc}
\toprule
			&\multicolumn{4}{c}{\textbf{Decoder layers}}\\
			\cmidrule(lr){2-5}
			 & 2& 4 & 6 & 8 \\
			\midrule
		One-shot & 49.4 & 51.8 & \colorbox{lightgray}{53.2} & 53.6\\
		Five-shot & 53.1 & 55.7 & \colorbox{lightgray}{56.6} & 56.8\\
		\bottomrule
\end{tabular}
}
}
\end{table*}

\textbf{Benefit of positional encoding.} We report results in Table~\ref{tab:positoinan encoding}. Including the spatio-temporal positional encoding is always beneficial, ideally in both the support and query encoder. Not passing any spatio-temporal positional encoding to the features leads to an mAP drop of 3.8 under the one-shot and an mAP drop of 4.3 under the five-shot setting.

\begin{table}
	\centering
	\caption{\textbf{Positional encoding analysis.} Adding a positional encoding for both the query and the support yields best performances }
	\label{tab:positoinan encoding}
	\resizebox{0.5\columnwidth}{!}{%
	\begin{tabular}{cccc}
	\toprule
	\multicolumn{2}{c}{\textbf{Positional encoding}} & \multicolumn{2}{c}{\textbf{Few-shot setting}}\\
	\cmidrule(lr){1-2} \cmidrule(lr){3-4}	
	support & query & one-shot & five-shot\\
	\midrule
	& & 49.4 & 52.3\\
    & $\checkmark$ & 50.3 & 53.7\\
    $\checkmark$ & & 52.4 & 55.9\\
	$\checkmark$ & $\checkmark$ & \textbf{53.2} & \textbf{56.6}\\
	\bottomrule
    \end{tabular}%
	}
\end{table}

\textbf{Transformer ablation.} In the Table~\ref{tab:self_ablation_module} of the paper, we show the effect of the three components in our proposed transformer under self-shot supports. Here we show in Table~\ref{tab:ablation_module} the results of the transformer ablation under few-shot setting with oracle supports and observe that the proposed transformer is also effective for video instance segmentation with few-shot oracle supports.
\begin{table}[t!]
	\centering
	\caption{\textbf{VIS transformer ablation with few-shot oracle supports.} Besides the transformer ablation under the self-shot setting reported in the paper (Table~\ref{tab:self_ablation_module}), we also report the effectiveness of the three modules in the proposed transformer under the few-shot setting with oracle supports
}
	\label{tab:ablation_module}
	\resizebox{0.6\columnwidth}{!}{%
	\begin{tabular}{ccccccc}
	\toprule
	\multirow{2}*{\textbf{Encoder}} & \multirow{2}*{\textbf{Fuser}} & \multirow{2}*{\textbf{Decoder}} & \multicolumn{2}{c}{\textbf{Self-VIS}} & \multicolumn{2}{c}{\textbf{Self-OVIS}}\\
		\cmidrule(lr){4-5} \cmidrule(lr){6-7}
	  &  &  & 1-shot & 5-shot & 1-shot & 5-shot\\
	\midrule
	& & $\checkmark$ & 42.9 & 43.4 & 15.0 & 15.4\\
    $\checkmark$ & & $\checkmark$ & 44.5 & 44.8 & 16.2 & 17.1\\
    & $\checkmark$ & $\checkmark$ & 50.6 & 53.7 & 20.9 & 22.3\\
	$\checkmark$ & $\checkmark$ & $\checkmark$ & \textbf{53.2} & \textbf{56.6}& \textbf{22.0} & \textbf{24.9}\\
	\bottomrule
    \end{tabular}%
	}
\end{table}

\textbf{Comparison against related methods.} As the task of few-shot VIS is novel, we cannot compare with existing methods that are not designed for this setting. We nevertheless try and thus adapt existing methods intended for other tasks. Michaelis~\etal~\cite{michaelis2018one} provides a strong baseline for few-shot image instance segmentation based on Mask R-CNN~\cite{he2017mask} (FMRCNN). 
They generate many segmentation proposals for the query image and match the proposals with the support image to find the fitting ones.  
To build a baseline for few-shot VIS, we first extract the frame features for the support and query videos by our backbone. Then we compute the average of the support frame features as in prototypical networks~\cite{snell2017prototypical}. 
Finally, with the averaged support feature, the instance segmenter can predict the instance masks for each query frame.
We also compare against Yang~\etal~\cite{yang2021few}, which is a transformer-based approach for few-shot common action detection (FST). 
They propose a dedicated encoder-decoder structure to learn the commonality and predict a spatio-temporal localization. 
We transplant their transformer structure in our pipeline by replacing our common transformer. We compare to both baselines under one- and five-shot settings on Self-VIS and Self-OVIS in Table~\ref{tab:comparison_vis}. Our approach achieves the best results with large margins on both datasets and settings.

\begin{table}
	\centering
	\caption{\textbf{Video instance segmentation} on Self-VIS and Self-OVIS. We constructed two baselines~\cite{michaelis2018one,yang2021few} from author-provided code. Our common transformer obtains favourable results under one- and five- shot settings on both datasets}
	\label{tab:comparison_vis}
	\resizebox{0.7\columnwidth}{!}{%
		\begin{tabular}{lrrrr}
			\toprule
		&\multicolumn{2}{c}{\textbf{Self-VIS}}&\multicolumn{2}{c}{\textbf{Self-OVIS}}\\
			\cmidrule(lr){2-3} \cmidrule(lr){4-5}
			& one-shot & five-shot & one-shot & five-shot\\
			\midrule
		SMRCNN~\cite{michaelis2018one} & 43.2 & 44.7 & 16.8 & 17.7\\
		FST~\cite{yang2021few} & 48.5 & 51.7 & 20.5 & 22.3\\
			\textit{\textbf{This paper}} & \textbf{53.2} & \textbf{56.6} & \textbf{22.0} & \textbf{24.9}\\	
			\bottomrule
		\end{tabular}%
	}

\end{table}
\begin{table}[t]
	\centering
	\caption{\textbf{Generalisation to images.} We report common instance segmentation performance on the video and images when only image support input is provided. 
	For video evaluation we use Self-VIS and for image evaluation MS-COCO from~\cite{michaelis2018one}. For MS-COCO we evaluate mAP at IoU level of $0.5$ as in~\cite{michaelis2018one} }
	\label{tab:comparison_vis_image}
		\resizebox{0.7\columnwidth}{!}{%
		\begin{tabular}{l rr rr}
			\toprule
				&\multicolumn{2}{c}{\textbf{img$\rightarrow$vid}} &\multicolumn{2}{c}{\textbf{img$\rightarrow$img}}\\
			\cmidrule(lr){2-3} \cmidrule(lr){4-5}
			& one-shot & five-shot & one-shot & five-shot\\
			\midrule
		SMRCNN~\cite{michaelis2018one} & 41.8          & 44.1         & 14.2 & 16.3 \\
		FST~\cite{yang2021few}         & 43.3          & 46.4         & 15.1 & 17.7\\
		\textit{\textbf{This paper}}         & \textbf{48.7} & \textbf{52.0} & \textbf{15.6} & \textbf{18.4}\\	
			\bottomrule
		\end{tabular}%
			}
\end{table}


\textbf{Video instance segmentation from images.} Besides using videos, we can also perform few-shot video instance segmentation using only images as support.
This is a novel and more challenging setting, as the model cannot exploit temporal information from the support videos. 
In Table~\ref{tab:comparison_vis_image}, (img$\rightarrow$vid), we perform few-shot video instance segmentation task with support images that are only a single middle frame for each support video using the Self-VIS dataset. 
We again compare against other methods and find that our approach performs better under both the one- and five- shot setting on this new task.
Extending this task even further, we also consider the problem of few-shot image instance segmentation:
Given a few support images containing the same object instances, find and segment the common object instances in a query image. 
As shown in Table~\ref{tab:comparison_vis_image}, (img$\rightarrow$img), our method can even generalize to this task on the challenging COCO dataset~\cite{michaelis2018one,lin2014microsoft} where the backbone network is pretrained on ImageNet dataset,
outperforming the other methods.

\subsection{Self-shot experiments}

\textbf{Self-shot versus zero-shot learning.} As a supplement to the experiments in Table~\textcolor{red}{2} in the paper, we construct another two zero-shot baselines by randomly cropping segments in time or in space from the query video as supports. Results in Table~\ref{zero_shot_segments} show that self-shot supports from a unlabelled pool of videos are more effective than the supports from the query video itself.
This demonstrates that sample diversity in the support helps improve performance.

\begin{table}[h] 
\footnotesize
        \centering
        \caption{\textbf{Self-shot versus zero-shot learning} for video instance segmentation and temporal action localization. The experiment setups follow Table~\textcolor{red}{2} in the paper.  We build two simple zero-shot baselines by adapting the temporal or spatial segments of the query video as supports. We find that self-shot supports are more effective than segments from the query video itself }
        \label{zero_shot_segments}
        \begin{tabular}{l c cc c cc}
        \toprule
        &&      \multicolumn{2}{c}{\textbf{Self-VIS}} && \multicolumn{2}{c}{\textbf{Thumos14}} \\
        \cmidrule{3-4} \cmidrule{6-7}
        && 1 & 5 && 1 & 5 \\
        \midrule
        Zero-shot learning (temporal segments) && 44.0 & 46.6 && 40.7 & 41.9 \\
        Zero-shot learning (spatial segments) && 44.9 & 47.8    && 41.1 & 42.4  \\
        Self-shot learning                       && 51.4  & 54.3  && 45.8 & 47.3 \\
        Self-shot learning k+(5)                 && 54.6  & 55.4  && 47.7 & 48.0 \\
        \bottomrule
        \end{tabular}
\end{table}


%
%
\bibliographystyle{splncs04}
\end{document}